\documentclass[preprint,10pt]{elsarticle}
\makeatletter
 \def\ps@pprintTitle{%
   \let\@oddhead\@empty
   \let\@evenhead\@empty
   \def\@oddfoot{\reset@font\hfil\thepage\hfil}
   \let\@evenfoot\@oddfoot
 }
\makeatother

\usepackage{amssymb}
\usepackage{amsmath}

\usepackage{tabularx}
\usepackage{graphicx}
\usepackage{comment}
\usepackage{float}
\usepackage{multirow}
\usepackage{array}
\usepackage{amsmath,amssymb,exscale}
\usepackage{lscape}
\usepackage[english,ruled,linesnumbered,algo2e]{algorithm2e}
\usepackage{enumitem}
\usepackage{subcaption}
\usepackage{hyperref}

\begin{document}

\begin{frontmatter}



\title{Enhancing Essay Cohesion Assessment: A Novel Item Response Theory Approach}


\author[label1,label6]{Bruno Alexandre Rosa} 

\affiliation[label1]{organization={CESAR School},
            city={Recife},
            state={PE},
            country={Brazil}}
\ead{bralexandre11@gmail.com}
\cortext[cor1]{Corresponding author}

\author[label2]{Hilário Oliveira}
\affiliation[label2]{%
  organization={Federal Institute of Espírito Santo},
  city={Serra},
  state={ES},
  country={Brazil}
}

\author[label3]{Luiz Rodrigues}
\affiliation[label3]{%
  organization={Computing Institute - Federal University of Alagoas},
  city={Maceió, Brazil},
  state={AL},
  country={Brazil}
}

\author[label6]{Eduardo Araujo Oliveira}
\affiliation[label6]{%
  organization={University of Melbourne},
  city={Melbourne},
  state={Victoria},
  country={Australia}
}
\author[label1,label4,label5]{Rafael Ferreira Mello}
\affiliation[label4]{%
  organization={AiBox Lab, Computing Department, Federal Rural University of Pernambuco},
  city={Recife},
  state={PE},
  country={Brazil}
}
            
\affiliation[label5]{%
organization={Centre for Learning Analytics, Monash University},
  city={Melbourne},
  state={Victoria},
  country={Australia}
}

\begin{abstract}
Essays are considered a valuable mechanism for evaluating learning outcomes in writing. Textual cohesion is an essential characteristic of a text, as it facilitates the establishment of meaning between its parts. Automatically scoring cohesion in essays presents a challenge in the field of educational artificial intelligence. The machine learning algorithms used to evaluate texts generally do not consider the individual characteristics of the instances that comprise the analysed corpus. In this meaning, item response theory can be adapted to the context of machine learning, characterising the ability, difficulty and discrimination of the models used. This work proposes and analyses the performance of a cohesion score prediction approach based on item response theory to adjust the scores generated by machine learning models. In this study, the \textit{corpus} selected for the experiments consisted of the extended Essay-BR, which includes 6,563 essays in the style of the National High School Exam (ENEM), and the Brazilian Portuguese Narrative Essays, comprising 1,235 essays written by 5th to 9th grade students from public schools. We extracted 325 linguistic features and treated the problem as a machine learning regression task. The experimental results indicate that the proposed approach outperforms conventional machine learning models and ensemble methods in several evaluation metrics. This research explores a potential approach for improving the automatic evaluation of cohesion in educational essays \footnote{Draft version 11/07/2025. This manuscript is a preprint and has not been peer-reviewed}.
\end{abstract}





\end{frontmatter}


\section{Introduction}\label{sec1}
Writing is a fundamental skill for academic, corporate and social development \citep{graham2019changing}. The essay, a common form of written assessment in educational settings, is frequently used to evaluate students' writing abilities and should be structured according to a clearly defined framework \citep{graham2019changing}. Due to its ability to assess writing proficiency, the essay is often included in entrance exams to select future students from various schools and universities \citep{Klein2009}. Producing an essay requires the appropriate use of linguistic mechanisms, which are essential for the development of writing skills \citep{travaglia2018tipologia}.

Assessing essays, however, is a subjective and time-consuming task, often requiring significant expertise to evaluate complex aspects like cohesion \citep{graham2019changing}. The evaluation process requires a careful analysis of how ideas are structured and interconnected, which varies widely depending on the writer’s style, intent, and audience. For instance, identifying cohesive elements involves interpreting grammatical links, semantic relationships, and logical flow across sentences and paragraphs. This requires not only an understanding of linguistic features but also an awareness of the broader context and purpose of the text. Furthermore, the inherently qualitative nature of cohesion assessment introduces subjectivity, as evaluators may differ in their judgments about the effectiveness of a text’s structure and clarity. The task becomes even more time-intensive when multiple essays need to be reviewed consistently, as maintaining fairness and reliability across assessments is challenging. These factors underscore the need for significant expertise and effort, which automated methods aim to alleviate by offering tools to assist educators and students -- at scale and consistently -- in this intricate process \citep{ferreira2024words}.

Despite their potential, current automated methods for assessing cohesion face noteworthy limitations. For instance, they often fail to capture the reference and sequence of semantic relationships, as well as the logical interconnection between the different parts of a text. In addition, cohesive elements often rely on contextual nuances that automated methods cannot fully replicate. These elements may carry multiple meanings depending on the context in which they are used, adding another layer of complexity. As a result, the automated assessment of cohesion in essays remains an open problem and a focus of ongoing research\citep{crossley2019tool,oliveira2022estimando,oliveira2023towards}.

One approach to addressing this challenge is to create a committee of classifiers tailored to perform scoring in specific contexts. Recent studies have taken a different perspective on the evaluation, assessing the performance of the algorithm at the instance level \citep{martinez2019item,moraes2022evaluating,uto2023integration}. In these specific cases, the main goal is to identify which instances in a dataset are more or less difficult for a given set of Machine Learning (ML) algorithms, allowing for a more precise and detailed analysis of the quality of predictions. This makes it possible to select more appropriate test sets and optimise the performance of ML algorithms, which can be useful for different purposes, both for learning and assessment. 

In this context, Item Response Theory (IRT) serves as a statistical framework for modelling the probability of a respondent answering an item correctly based on their latent ability \citep{embretson2013item}. When applied to evaluating ML algorithms, IRT can model the prediction responses of the algorithms in relation to the characteristics of the instances \citep{martinez2019item}. This enables the identification of the most informative instances for enhancing prediction performance. Recent studies have produced promising results in this context \citep{moraes2022evaluating,uto2023integration}. These studies compare the application of IRT across different regression models and provide a theoretical analysis of its parameters and capabilities in ML models and contexts. However, further research is needed to explore how IRT can be leveraged to adjust and combine the predictions of AI algorithms, paving the way for a novel approach to generating cohesion scores.

In this study, IRT is employed to adjust and combine the output predictions of AI algorithms, offering a novel method for predicting cohesion scores in essays. This approach was tested using the extended Essay-BR dataset \citep{marinho2022automated}, which consists of 6,563 essays written in the style of the National High School Exam (ENEM), and the Brazilian Portuguese Narrative Essays dataset \citep{mello2024propor, OliveiraCorpusNarrativeEssays2025}, containing 1,235 essays written by 5th to 9th grade public school students. By extracting 325 linguistic features -- spanning lexical, syntactic, and semantic properties -- and framing the task as a machine learning regression problem, the approach demonstrated superior results than traditional ML algorithms (such as XGBoost, Support Vector Machine, Multilayer Perceptron and others), ensemble techniques (e.g., Voting and Stacked), and Bidirectional Encoder Representations for Transformers (BERT)-based models. These results indicate that the IRT-based method provides a more precise automated evaluation of textual cohesion in educational essays, with significant implications for the development of advanced Automated Essay Scoring (AES) systems.

\section{Background and Related Works}\label{sec:background}

\subsection{Cohesion in Student Essays}\label{sec:cohesion}

Cohesion is an essential aspect of a text, as it establishes meaningful relationships between its components. \cite{halliday1976cohesion} defines cohesion as the property that creates and signals any connections and links, providing a text with semantic unity. Similarly, \cite{koch2010coesao} describes textual cohesion as the use of vocabulary and grammatical structures that connect ideas within a text.  Cohesion functions are used to create, establish and signal the links that link and articulate different segments of a text, ensuring its interpretability.

Researchers have extensively studied the relationship between computationally extracted writing features and human cohesion assessments. For example, \cite{crossley2010cohesion} found statistically significant negative correlations between human ratings of cohesion and features such as anaphoric references, measures of causal cohesion, frequency of connectives, and overlap measures. Using the Coh-Metrix tool, the study identified three highly predictive features: syntactic complexity (number of words before the main verb), lexical diversity, and word frequency. 

\cite{crossley2019tool}  extended this work by investigating paragraph-level cohesion. They found that overlap between adjacent paragraphs positively correlated with human cohesion ratings, while verb overlap between sentences was a negative predictor. Their development of the TAACO tool introduced several indices related to local and global cohesion at the semantic level. The findings suggested that these indices significantly predict text cohesion and speech quality, advancing the understanding of cohesive features and their relationship to human assessments in English texts.

The analysis of textual cohesion in Brazilian Portuguese has been less explored. \cite{lima2018automatic} investigated 91 features, encompassing dimensions of lexical diversity, connective count, readability indices, and word overlap measures between sentences and paragraphs. An analysis of the performance of a support vector machine classification model was conducted using a corpus of 6,867 essays written by high school students. The performance was 0.58 for Precision and 0.26 for Recall. The study highlighted the scarcity of data and tools for Portuguese, emphasising the need for further research in this area. 

To address this gap, \citet{grama2022elementos} developed a tool called CoTex to assist high school students in improving their use of cohesive elements in essays. CoTex incorporates features such as a dictionary-like word list, identification of sense relations, exercises for practicing cohesive element usage, and a diversity analysis tool. Although CoTex supports students in recognising and using cohesive elements, it does not employ ML techniques to automate the cohesion correction process. Instead, it relies on a manually curated dictionary.

Further advancements in automated cohesion analysis in Portuguese include studies by \cite{oliveira2022estimando} and \cite{oliveira2023towards}. \cite{oliveira2022estimando} performed an analysis considering 151 features that encompass aspects such as the use of connectors, lexical diversity, readability, and similarity between adjacent sentences, as well as several features extracted from the Coh-Metrix tool. The study compared several regression algorithms to estimate the cohesion score using the Essay-BR database \citep{marinho2022automated}. Building on this, \cite{oliveira2023towards} investigated various regression-based approaches, from feature-based methods to the application of BERT-based models, to estimate scores related to textual cohesion in Portuguese and English datasets. Their findings showed that BERT-based models outperformed traditional ML approaches. For example, in the Essay-BR dataset, the BERT-based model with 20 epochs achieved an RMSE of 0.164, while in the ASAP++ dataset, the BERT-based model with one epoch achieved an RMSE of 0.149. Explainability methods were also employed to interpret model decisions.

Despite these advancements, most automated cohesion analysis studies focus on the English language. Tools such as Coh-Metrix \cite{crossley2010cohesion} and TAACO \cite{crossley2019tool} have proven effective in correlating text features with human cohesion assessments. However, there remains a significant gap in the literature for other languages, particularly Portuguese. Studies such as \cite{lima2018automatic,grama2022elementos,oliveira2022estimando,oliveira2023towards} have initiated efforts to address this gap, but much remains to be explored.  With recent advances in statistical modelling techniques and machine learning, new opportunities have emerged for researchers to investigate and improve predictions of textual cohesion.

\subsection{Item Response Theory in Machine Learning}\label{sec:background_irt}

IRT is a statistical framework widely used in psychometrics to estimate human ability based on responses to test items with varying levels of difficulty \cite{embretson2013item}. This approach is commonly used in large-scale educational exams to estimate student latent ability and the characteristics of questions based on parameters such as difficulty, discrimination, and/or guessing. A notable advantage of IRT is its ability to assess diverse skills of individuals with similar test performances by considering item-specific parameters \citep{embretson2013item}. Additionally, IRT allows item difficulty to be calibrated relative to the average level of the population's latent construct, enabling adaptation of items for different populations, thus increasing the accuracy and reliability of assessment \citep{embretson2013item}.

In IRT, an individual's ability is estimated on the basis of their responses to test items. This ability is mapped to the probability of correctly answering an item through mathematical functions that generate the Item Characteristic Curve (ICC) \citep{embretson2013item}. Recently, researchers have explored the application of IRT models for instance-level evaluations in ML\citep{moraes2022evaluating,uto2023integration,oldfield2024itemresponsetheorybasedr}. In this context, ML models are treated as "respondents", while individual instances are considered ``test items'', allowing the parameters estimated by IRT to identify which instances are more challenging in a dataset.

\cite{uto2023integration} proposed an IRT-based method to assess the characteristics of scores estimated by machine learning algorithms in AES and to integrate these results into a final estimated score. The study demonstrated that the proposed method, using the IRT Generalized Many-Facet Rasch Model (GMFRM), achieved superior average precision compared to both individual AES models based on machine learning. For example, the SkipFlow AES model achieved a Quadratic Weighted Kappa (QWK) of 0.7209, while STACKING, an ensemble integration method, achieved a QWK of 0.7395. The proposed GMFRM model outperformed both, achieving a QWK of 0.7562. These results highlight that the integration of multiple AES models using IRT can significantly improve performance, evidencing its potential. However, the study was limited to testing a restricted set of AES and IRT models, did not perform a cross-validation analysis and only applied the method to one English corpus. More research is needed to evaluate the broader applicability of IRT to integrate AES models.

\cite{chen2019beta} and \cite{ferreira2023beta} explored another application of IRT in ML by modelling absolute error as a function of IRT parameters using a Beta distribution ($\beta$). This approach generates an error expectation adapted for continuous positive-valued responses. Their work demonstrated how IRT parameters could identify regions of high and low difficulty within a dataset, influenced by complex patterns in the data or noise. The core equations of the model, including the discrimination and difficulty parameters, are defined by Equation \ref{eq_model_irt}. 

\begin{equation}
\begin{aligned}
\mathbb{E}\left[p_{ij}\mid\theta_i,\delta_j,\omega_j,\tau_j\right] = \frac{1}{1 + 
\left( \frac{\delta_j}{1-\delta_j} \right)^{\tau_j \cdot \omega_j}
\cdot \left( \frac{\theta_i}{1-\theta_i} \right)^{-\tau_j \cdot \omega_j}}
\end{aligned}
\label{eq_model_irt}
\end{equation}

In Equation \ref{eq_model_irt}, $\mathbb{E}$ is the probability of respondent $i$, with the ability of respondent $\theta_i$, responding correctly to item $j$. The item parameters include Discrimination ($\tau_j \cdot \omega_j$), representing how much the item $j$ differentiates between high-quality and low-quality responses. The higher its value, the more discriminating the item; Difficulty ($\delta_j$), which represents how difficult it is for an item to be answered correctly, and the higher its value, the more difficult the item.

These studies \cite{ferreira2023beta,uto2023integration} demonstrate advances in the application of IRT in ML algorithms, each approached from a different perspective. The study by \cite{ferreira2023beta} compares the use of IRT in different regressors and provides a theoretical analysis of its parameters and abilities in ML models. The research by \cite{uto2023integration} focused on integrating the results of AES using IRT. However, unlike previous studies, the present research employs IRT to adjust the predictions of ML algorithms and proposes a novel approach for predicting cohesion scores in Brazilian Portuguese essays.

\section{Method}

The goal of this study is to investigate the viability of using IRT to estimate human ratings of essay cohesion scores in Portuguese. These scores were generated by traditional ML algorithms, ensemble methods, and BERT-based deep learning models. To achieve this, we extracted a comprehensive set of linguistic features based on prior theoretical and empirical works \citet{palma2018,Mello2022,oliveira2022estimando,oliveira2023towards}. These features were derived from students' essays written in Portuguese and were used to develop artificial intelligence models. Subsequently, IRT was applied to adjust the output predictions of these models, forming a novel approach to predicting essay cohesion scores. 

This study is guided by the following research question (RQ): To what extent can a model based on item response theory accurately adjust the cohesion scores of essays generated by artificial intelligence algorithms?

Our investigation was structured into five stages to systematically address our research question.

\subsection{Dataset Selection and Preprocessing}\label{sec:data}

The first dataset used was the Essay-BR, which comprises essays written in the style of the \textit{Exame Nacional do Ensino Médio} (ENEM), Brazil's most significant national assessment for secondary education \citep{Klein2009}. Established in 1998, ENEM evaluates students’ readiness at the end of basic education. In 2009, it evolved into a gateway for higher education admissions, attracting millions of candidatures annually. In 2023, more than 3.9 million candidates registered with ENEM \footnote{https://www.gov.br/inep/pt-br/assuntos/noticias/enem/3-9-milhoes-estao-inscritos-no-enem-2023} producing argumentative essays on various scientific, cultural, political or social topics \citep{Klein2009}.

The Essay-BR dataset, collected by \cite{marinho2022automated}, consists of 6,579 essays written between December 2015 and August 2021. These essays were sourced from public portals, including Vestibular UOL\footnote{https://vestibular.brasilescola.uol.com.br/banco-de-redacoes} and Educação UOL\footnote{https://educacao.uol.com.br/bancoderedacoes}, and covered 151 topics ranging from human rights and health issues to cultural activities and the COVID-19 pandemic. The dataset adheres to ENEM's essay format, requiring texts between 8 and 30 lines\footnote{https://www.gov.br/inep/pt-br/areas-de-atuacao/avaliacao-e-exames-educacionais/enem}.

Each essay was assessed on five competencies, with scores ranging from 0 to 200 for each competency. The competencies include:
\textbf{(i)} Command of the formal written form of Portuguese; 
\textbf{(ii)} Understanding of the proposed topic and application of interdisciplinary concepts to develop the theme within the structural limits of the essay in prose;
\textbf{(iii)} Ability to select, relate, organise and interpret information, facts, opinions and arguments to defend a point of view;
\textbf{(iv)} Demonstration of linguistic mechanisms for constructing arguments; and
\textbf{(v)} Development of a human rights-compliant intervention proposal.

The final scores ranged from 0 to 1,000 and were calculated as the sum of the competency scores. Essays were evaluated by two expert raters, with discrepancies exceeding 80 points resolved by a third assessor. The final score for each competency was the arithmetic mean of the two closest evaluations.


The second dataset comprises 1,235 narrative essays written by public school students in the 5th grade \citep{OliveiraCorpusNarrativeEssays2025}. This corpus was used in the PROPOR'24 competition \citep{mello2024propor}, which aimed to promote the development of automated essay evaluation systems in Portuguese. The essays were produced in classrooms based on predefined prompts provided by teachers. All texts were manually transcribed and anonymized to protect student identities.

The essays were evaluated according to a rubric established by the Brasil na Escola programme, with four main competencies: \textbf{(i)} Formal register; \textbf{(ii)} Thematic coherence; \textbf{(iii)} Textual typology; and \textbf{(iv)} Textual cohesion. For each competency, a score was assigned ranging from I (low proficiency) to V (complete proficiency). Two human assessors corrected the essays, with a third involved in assessing any discrepancies.
 
Table \ref{tab:statistics_essay} presents descriptive statistics about the datasets, including cohesion scores, the total number of essays, the mean of sentences, the mean of words, and the standard deviation (presented in parentheses). Essays in the Essay-BR dataset averaged 297.30 words and 11.22 sentences, while those in the Narrative Essays dataset averaged 177.85 words and 7.91 sentences.

\begin{table*}[!ht]
    \small
    \caption{Descriptive statistics of the Portuguese essay datasets}
    \label{tab:statistics_essay}
    \centering
    \begin{tabular}{lrrrr}
    \hline
    \multicolumn{1}{c}{\textbf{Dataset}} & \multicolumn{1}{c}{\textbf{\begin{tabular}[c]{@{}c@{}}Cohesion\\ Score\end{tabular}}} & \multicolumn{1}{c}{\textbf{\#Essay}} & \multicolumn{1}{c}{\textbf{\begin{tabular}[c]{@{}c@{}}Mean \\ Sentences\end{tabular}}} & \multicolumn{1}{c}{\textbf{\begin{tabular}[c]{@{}c@{}}Mean \\ Words\end{tabular}}} \\ \hline
    \multirow{7}{*}{\begin{tabular}[c]{@{}l@{}}Extend \\Essay-BR (EEBR) \\ \cite{marinho2022essay}\end{tabular}} & 0 & 206 & 8.01 (3.70) & 225.24 (67.89) \\
     & 40 & 65 & 8.35 (3.74) & 217.06 (87.98) \\
     & 80 & 879 & 9.42 (4.68) & 249.62 (85.34) \\
     & 120 & 2,455 & 10.69 (4.69) & 283.34 (79.26) \\
     & 160 & 1,821 & 12.06 (3.78) & 320.94 (72.18) \\
     & 200 & 1,137 & 13.15 (3.47) & 344.10 (67.94) \\
     & \textit{Overall} & \textit{6,563} & \textit{11.22 (4.42)} & \textit{297.30 (83.25)} \\ \hline
    \multirow{6}{*}{\begin{tabular}[c]{@{}l@{}}Brazilian Portuguese\\ Narrative Essays (BPNE) \\ \cite{OliveiraCorpusNarrativeEssays2025}\end{tabular}} & 1 & 40 & 6.35 (5.44) & 100.73 (61.67) \\
     & 2 & 187 & 7.01 (6.94) & 136.61 (70.81) \\
     & 3 & 824 & 7.95 (7.09) & 185.50 (80.73) \\
     & 4 & 157 & 8.45 (5.66) & 197.42 (72.32) \\
     & 5 & 27 & 12.26 (9.28) & 230.48 (90.84) \\ 
     & \textit{Overall} & \textit{1,235} & \textit{7.91 (6.96)} & \textit{177.85 (81.70)} \\ \hline
    \end{tabular}
\end{table*}

Both datasets exhibit significant imbalances in the distribution of cohesion scores. In the EEBR dataset, essays scored at 120 represent 37.41\% of the corpus (2,455), with scores of 160 and 200 comprising 27.75\% (1,821) and 17.32\% (1,137), respectively. In contrast, texts with scores of 0, 40, and 80 represent only 3.14\% (206 essays), 0.99\% (65) and 13. 39\% (879) of the total number of essays. Similarly, the analysis is present in BPNE dataset. Essays with a score of 3 represent 66.72\% of the corpus (824), followed by scores of 2 (15.14\%, 187) and 4 (12.71\%, 157). Scores 1 and 5 represent only 3.24\% (40) and 2.19\% (27). These imbalances in both datasets, especially in the lower and higher grades, present challenges for machine learning models, necessitating careful consideration during training and evaluation.

Although the cohesion scores in the datasets adopted for the study were discrete, we decided to explore regression models instead of classifiers to align with previous studies treating text cohesion analysis as a regression task \citep{lima2018automatic,grama2022elementos,oliveira2022estimando,oliveira2023towards}. This decision ensures consistency with previous approaches, including validation processes and performance measures \citep{marinho2022automated,oliveira2023towards}.

\subsection{Linguistic Feature Extraction}

\subsubsection{Feature-based Approach}\label{sec:features}

The purpose of this stage is to transform textual data into numerical vectors that can be processed by machine learning models while preserving the underlying meaning of the text. We computed a comprehensive set of language-independent features, commonly used in educational text mining and learning analytics research \citep{sha2021hammer}. These features have previously been used to analyse rhetorical structures in essays \citep{Mello2022}, automate essay scoring \citep{ramesh2022automated}, evaluate online discussions \citep{mello2021towards},  and assess textual cohesion \citep{crossley2019tool,oliveira2022estimando,oliveira2023towards}.

Table \ref{tab:features_group} summarises the 325 linguistic features used in this study. The selection was informed by prior research demonstrating their efficacy in essay evaluation, particularly in assessing textual cohesion. Although some resources cover overlapping features, variations in their implementation can lead to distinct outputs. Therefore, we utilised all available resources to ensure a comprehensive analysis. 
To manage the extensive range of indices, the features were grouped into 13 categories, as outlined in Table \ref{tab:features_group}. Each category represents a distinct linguistic dimension, offering a broad perspective on the characteristics of the essays. The following subsections provide detailed explanations of the 13 categories.

\begin{table*}[!htbp]
    \small
    \caption{Features per Group}
    \label{tab:features_group}
    \centering
    \begin{tabular}{@{}lccc@{}}
    \hline
    \multicolumn{1}{c}{\textbf{Group}}       & \multicolumn{1}{c}{\textbf{Reference}} & \multicolumn{1}{c}{\textbf{\#Feature}} & \multicolumn{1}{c}{\textbf{\%}}   \\ 
    \hline
Coh-Metrix                          & \cite{camelo2020coh}  & 87                                            & 26.77                          \\
LIWC                         & \cite{balage2013evaluation} & 64                                            & 19.69                          \\
Word Morphsyn Information            & \cite{leal2021nilc} & 39                                            & 12.00                          \\
Connectives                    & \cite{grama2022elementos} & 33                                            & 10.15                          \\
Semantic Cohesion                    & \cite{leal2021nilc} & 21                                            & 6.46                           \\
Lexical Diversity                    & \cite{leal2021nilc} & 18                                            & 5.54                           \\
Syntactic Complexity                 & \cite{leal2021nilc} & 17                                            & 5.23                           \\
Descriptive Measures                & \cite{leal2021nilc} & 11                                            & 3.38                           \\
Referential Cohesion                & \cite{leal2021nilc} & 9                                             & 2.77                           \\
Textual Simplicity                  & \cite{leal2021nilc} & 8                                             & 2.46                           \\
Sequential Cohesion                 & \cite{leal2021nilc} & 7                                             & 2.15                           \\
Readability                   & \cite{palma2018} & 7                                             & 2.15                           \\
Syntactic Patterns Density          & \cite{leal2021nilc} & 4                                             & 1.23                           \\ \hline
                     \textbf{Overall}           &     & \textbf{325}       & \textbf{100}\%         \\ \hline
    \end{tabular}  
\end{table*}

\begin{itemize}
    \item \textbf{Coh-Metrix:} Described in \cite{mcnamara2014automated}, it generates indices for linguistic and discourse representations of an essay. It is commonly used in the literature to extract indices from educational essays, including measures for cohesion analysis \citep{ramesh2022automated}. In this research, we employed 87 linguistic descriptors in the Portuguese version of Coh-Metrix \citep{camelo2020coh}. 
    \item \textbf{LIWC:} The Linguistic Inquiry Word Count (LIWC) dictionary \citep{pennebaker2001linguistic} calculates the degree of usage of different word categories in a wide range of texts. The Portuguese version of LIWC 2007 \citep{balage2013evaluation} refers to a lexicon with 127,149 words organised into 64 categories.
    \item \textbf{Word Morphsyn Information:} In this group of indices, various morphosyntactic aspects are considered based on \citep{leal2021nilc}. In total, 39 features were extracted. This analysis primarily considers statistics related to adjectives, adverbs, nouns, and verbs, observing metrics such as proportion, maximum values, minimum values, and standard deviation.
    \item \textbf{Connectives:} In this study, we used connectives such as conjunctions and prepositions because they are crucial in building a cohesive and coherent text \citep{halliday1976cohesion,koch2010coesao}. The grouping of cohesive elements, according to the type of sense relationship they establish, was based on literature classifications \citep{halliday1976cohesion,koch2010coesao}. In \cite{grama2022elementos}, the author provided a set of these connectives extracted from ENEM-style essays. A total of 33 features were identified.
    \item \textbf{Semantic Cohesion:} The semantic cohesion of an essay is crucial for comprehension and clarity of the text, reflecting the author's ability to maintain thematic relevance and continuity of ideas \citep{halliday1976cohesion,koch2010coesao}. Approaches based on Latent Semantic Analysis (LSA) allow for the evaluation of semantic proximity between text segments, providing cohesion indicators. In total, 21 features were extracted based on the study \cite{leal2021nilc}.
    \item \textbf{Lexical Diversity:} In the case of lexical diversity indices, a wide range of metrics is considered from the \cite{leal2021nilc}. Analysing lexical features in essays is essential for assessing a writer's richness and variety of vocabulary \cite{halliday1976cohesion,koch2010coesao}. In total, 18 lexical diversity features were extracted.
    \item \textbf{Syntactic Complexity:} In text assessment, utilising syntactic complexity analysis represents a significant pillar for understanding the author's writing skills. According to \cite{leal2021nilc}, such metrics examine the grammatical structure of the text, going beyond vocabulary to observe how words are effectively organised into sentences. In total, 17 features were extracted.
    \item \textbf{Descriptive:} Descriptive analysis of texts is essential to understanding the macro-structure of an essay. Quantitative elements, such as the total number of paragraphs and sentences, provide an overview of content organisation and segmentation. Evaluating the average number of sentences per paragraph allows inferences about thematic cohesion and the delineation of main ideas. In total, 11 features were extracted.
    \item \textbf{Referential Cohesion:} Referential cohesion in essays is a crucial aspect that influences both the readability and communicative effectiveness of the text \citep{halliday1976cohesion,koch2010coesao}. When analysed together, these indicators provide a comprehensive view of referential cohesion, which is fundamental for reader comprehension and the overall quality of the text. In total, nine features were extracted from \cite{leal2021nilc}.
    \item \textbf{Textual Simplicity:} Textual simplicity is an essential dimension in the analysis of essays, especially when aiming to identify clarity in writing. For example, the proportion of pronouns used in dialogues can indicate an attempt to connect with the reader or a more colloquial narrative structure. 
    In total, eight features were extracted from the study \cite{leal2021nilc}. 
    \item \textbf{Sequential Cohesion:} Sequential cohesion in an essay is a crucial aspect that pertains to how ideas and information are organised and connected in a logical progression \citep{halliday1976cohesion,koch2010coesao}.
    The use of local cohesion indices, calculated based on specific text units such as words and phrases, provides an understanding of cohesion within delimited segments of the text. In total, seven features were extracted from  \cite{leal2021nilc}.
    \item \textbf{Readability:} The readability of a text is a crucial dimension that determines how easily a reader can comprehend the presented content. To assess this aspect, various indices that consider elements such as word length, sentence complexity, and lexical density are applied. In total, seven features were extracted from \cite{palma2018}.
    \item \textbf{Syntactic Patterns Density:} The density of syntactic patterns in a text refers to the frequency and structure of specific grammatical constructions that shape and nuance the writing. In essays, this density can be analyzed through metrics that examine the usage and organisation of certain linguistic elements. In total, four features were extracted from \cite{leal2021nilc}.
\end{itemize}

\subsubsection{BERT-based Approach} \label{BERT}

In recent years, learning analytics researchers have increasingly utilised pre-trained language models such as BERT to generate features for various tasks, including classification and natural text generation \citep{Li2021PretrainedLM}. These models are trained on large collections of textual data and take advantage of the Transformer architecture \citep{Vaswani2017Transformer}, which has revolutionised natural language processing (NLP) by enabling state-of-the-art performance in multiple domains \citep{Li2021PretrainedLM}.

Building on these advancements, this study explored the viability of using the pre-trained language models alongside the traditional feature extraction approaches described earlier. Specifically, we used the BERT language model \citep{devlin2018bert}, which is designed to provide contextual embeddings by processing input in a bidirectional way.  This allows BERT to generate semantic representations that are sensitive to variations in meaning based on the context in which words appear \citep{devlin2018bert}.

Using BERT, we obtained contextual word representations from our essay datasets in the form of embeddings. These embeddings capture the relationships between words and their contexts, enhancing the ability to model textual cohesion. The extracted embeddings were then used as input features to a feed-forward neural network, which was trained to estimate cohesion scores for the essays. This approach complements traditional linguistic features by leveraging deep contextual information, offering a more comprehensive framework for evaluating textual cohesion.

\subsection{Predictive Model Development and Training}

In this stage, the aim is to select, train, validate, and test different artificial intelligence models to estimate cohesion scores in essays. Although the cohesion scores in the datasets adopted for the study are discrete (see Table \ref{tab:statistics_essay}), it was decided to explore regression models instead of classifiers. This decision was made on the basis of the literature on analysing text cohesion and even automated essay scoring systems, such as regression problems \citep{Rosa2024CombinaoDM,oliveira2022estimando,oliveira2023towards}.

For this task, we trained different regression algorithms designed to predict cohesion scores of essays written in Portuguese. The selected algorithms, drawn from prior research \cite{ferreira2019text,oliveira2023towards}, represent diverse approaches, including statistical models, decision trees, classical neural networks, Bayesian models, and ensemble techniques. The ML algorithms employed in the study were:
\textit{Bayesian Ridge}\footnote{\label{scikit-learn}https://scikit-learn.org/}, 
\textit{CatBoost Regressor}\footnote{https://catboost.ai/},
\textit{Decision Tree Regressor}\footref{scikit-learn}, 
\textit{Extra Trees Regressor}\footref{scikit-learn}, 
\textit{LGBM Regressor}\footnote{https://github.com/Microsoft/LightGBM/}, 
\textit{Linear Regression}\footref{scikit-learn}, 
\textit{MLP Regressor}\footref{scikit-learn}, 
\textit{Random Forest}\footref{scikit-learn}, 
\textit{SVR}\footref{scikit-learn} e 
\textit{XGB Regressor}\footnote{https://github.com/dmlc/xgboost/}.
For all algorithms, the default hyperparameters provided by the respective libraries were used to streamline the training process.

To improve prediction performance, we also implemented ensemble methods, which combine the output of multiple models to produce more robust predictions \citep{sagi2018ensemble}. Two popular ensemble techniques were evaluated: (i) Voting Regressor, which averages the individual predictions of the regressors to form a final prediction, and (ii) Stacked Regressor, which combines predictions from multiple base regressors using a final meta-regressor. In this study, we used Linear Regression and SVR as meta-regressors for the stacked ensemble.

In addition to traditional and ensemble algorithms, the BERT neural language model was used in conjunction with an additional linear dense layer to estimate textual cohesion in essays. BERT represents one of the most significant NLP advances in automated essay scoring \citep{ferreira2024words}. This approach has been investigated for its ability to capture not only the isolated meanings of words, but also deep semantic relationships in texts. Following the guidelines of the work \cite{oliveira2023sbie}, we used the BERTimbau-base implementation \citep{souza2020bertimbau} and a compressed version of this model, named DistilBERT\footnote{https://huggingface.co/adalbertojunior/distilbert-portuguese-cased}. In line with the previous study, the fine-tuning process was carried out with the configuration of five training epochs.

By leveraging these diverse algorithms and ensemble strategies, it was possible to identify the most effective models for predicting cohesion scores, setting a foundation for further refinements. 

\subsection{IRT-Based Score Adjustment}

The main novelty of the current study lies in adopting IRT concepts to combine the cohesion scores predicted by the ML models to generate a final score. IRT provides a framework for modelling and understanding the interaction between individuals (respondents) and test items. In the context of machine learning, IRT can be adapted by treating instances in the dataset as items and ML algorithms as respondents \citep{moraes2022evaluating, uto2023integration}. The proposed approach integrates IRT into the score adjustment process, leveraging the BIRT-GD framework \citep{ferreira2023beta} to calculate error expectations. The model is particularly suitable because of its ability to deal with the symmetry of IRT parameters, using the gradient descent method for estimation. This allows for more accurate modelling and provides the flexibility to adjust for the characteristic curves of items with different formats.

The parameters of the IRT models described in Section \ref{sec:background_irt} can also have their concepts associated with ML models. Ability represents the capacity of each ML model to correctly predict cohesion scores. Difficulty indicates the complexity of predicting cohesion scores, reflecting the diversity and precision of the cohesive elements presented in the essays. And discrimination associates the sensitivity of a model to differentiate between essays at different levels of cohesion. Algorithm \ref{alg-tri} summarised the main steps investigated in this study. In general terms, the proposed approach, named IRT-Multiregressor, adjusts the predictions of ML models, reducing individual algorithm biases and aligning the results with established rubrics.

\begin{algorithm2e}[!hbt]
\caption{Pseudo-code of proposed IRT-Multiregressor approach}\label{alg-tri}
\KwData{essays $X$, human scores $y$, models $m$}
\KwResult{final score $FS$}

\For{$idx\_train, idx\_test \in sk.split(X)$}{

    $X_{train} \gets X[idx\_train]$, \quad $y_{train} \gets y[idx\_train]$
    
    $X_{test} \gets X[idx\_test]$, \quad $y_{test} \gets y[idx\_test]$

    $X_{train}, X_{valid}, y_{train}, y_{valid} \gets$ split($X_{train}, y_{train}$)

    $m_{train} \gets$ train($m, X_{train}, y_{train}$)

    $y_{valid}^{pred} \gets$ predict($m_{train}, X_{valid}$)

    $e_{ij} \gets |y_{valid} - y_{valid}^{pred}|$

    \For{$i \in Rubric$}{
        $\mathbb{E}_i \gets$ train\_birt-gd($e_{ij}, i$)
        
        \For{$m \in m_{train}$}{
            $L_{min}[i][m], L_{max}[i][m] \gets$ $[y_{valid}^{pred} - \mathbb{E}_i, y_{valid}^{pred} + \mathbb{E}_i]$
            
        }
    }

    \For{$r \in X_{test}$}{
        \For{$i \in Rubric$}{
            \For{$m \in m_{train}$}{
                \If{$y_{test}^{pred}[m] \in [L_{min}[i][m], L_{max}[i][m]]$}{
                    $b_m[i] \gets 1$
                }
                \Else{
                    $b_m[i] \gets 0$
                }
            }

            $recurrence\_interval[i] \gets$ count\_m($b_m[i]$)
            
            \If{tie($recurrence[i]$)}{
                $min_{error} \gets$ lowest\_error($recurrence, y_{test}^{pred}, i$)
            }
        }

        $FS \gets$ FS($recurrence, min_{error}$)
    }
    
    evaluate($FS, y$)
}
\end{algorithm2e}

As seen in Algorithm \ref{alg-tri}, the ML models made their respective predictions, generating a results matrix. Based on these predictions, the validation data set was used as the basis for calculating the intrinsic parameters of the IRT. Then, the BIRT-GD model receives a set of data, where $X$ is the matrix containing the absolute error ($e_{ij}$) generated between the human evaluator's score and the score predicted by each AI model. The index ($i$) refers to the item, which corresponds to the cohesion score range. The index column ($j$) represents the respondent, which refers to each ML model used. For each ML model and cohesion score range, the error expectation provides an expected error value, which is used to define the minimum and maximum limits of the predictions. These confidence intervals are essential for selecting the final prediction of the cohesion score, as they make it possible to identify the most frequent and reliable prediction among the models.

\subsection{Model Evaluation and Validation}

In this final stage, the performance of the results of different algorithms, ensembles, and the proposed approach were evaluated. We adopted a 10-fold stratified cross-value methodology to ensure the consistency and integrity of analysis results. In addition, 10\% of the training set in each of the ten executions was separated as the validation set. The following evaluation measures were used: linear Kappa coefficient \citep{cohen1960coefficient}, quadratic weighted Kappa \citep{cohen1968weighted}, Pearson's correlation \citep{pearson1896vii} and confusion matrix.

We selected these assessment measures because they are commonly used in the literature to evaluate automated essay scoring systems \citep{lima2018automatic,oliveira2022estimando,marinho2022automated,oliveira2023towards,de2023avaliaccao}. Although the Kappa, QWK, and confusion matrix are coefficients traditionally used in classification problems, in this context, they also apply to regression problems because the values predicted by the regressors were categorised into predefined score ranges predefined in the large-scale exam guidelines. The \textit{Kappa} and QWK metrics were calculated using the Scikit-learn\footnote{https://scikit-learn.org/} library, and the \textit{Pearson} correlation was calculated using the SciPy\footnote{https://scipy.org/} library.

Considering that the outputs from the regression algorithms are continuous, we applied the conversion scale delineated in previous research \cite{marinho2022automated}.  Values are mapped to scores in bands: values below 20 receive a score of 0; values from 20 to less than 60 receive a score of 40; from 60 to less than 100, a score of 80; from 100 to less than 140, a score of 120; from 140 to less than 180, a score of 160; and values of 180 or above receive a score of 200.

The Kappa is interpreted following the guidelines presented by \cite{landis1977measurement}, which indicate: (i) Values less than 0.20 suggest a low level of agreement; (ii) Values between 0.21 and 0.4 indicate a fair level of agreement; (iii) Values between 0.41 and 0.6 represent a moderate level of agreement; (iv) Values between 0.61 and 0.8 suggest a good level of agreement; (v) Values between 0.81 and 1.0 indicate a very high level of agreement.

We adopted the criteria outlined in \cite{ratner2009correlation} to interpret the Pearson correlation values, according to which: (i) a value of 0 indicates the absence of a linear relationship; (ii) values of +1 or -1 denote a perfect positive or negative relationship, respectively; (iii) scores ranging from 0 to 0.3 (or 0 to -0.3) correspond to a weak positive (or negative) association; (iv) scores between 0.3 and 0.7 (or -0.3 and -0.7) reflect a moderate positive or negative relationship; and (v) values between 0.7 and 1.0 (or -0.7 and -1.0) are interpreted as strong positive or negative correlations.

\section{Results}\label{sec:results}

The metrics were evaluated in experiments on two datasets to answer the RQ. The evaluation metrics were Kappa, Quadratic Weighted Kappa (QWK), and Pearson correlation. The results were presented separately for the extended Essay-BR (EEBR) and Brazilian Portuguese Narrative Essays (BPNE) datasets. The standard deviations were presented in parentheses. The best performing results for each metric and approach were emphasised in bold, and the best overall with the symbol (†).

Table \ref{tab:result_extend_essay} presents the performance of different approaches evaluated on the first EEBR dataset. Among the traditional ML models, SVR obtained the highest Kappa (0.250), while Cat Boost Regressor had the highest QWK (0.533) and Pearson correlation (0.572). Regarding the ensemble approaches, Stacked Linear Regression achieved the best results 0.253, 0.538, and 0.580, respectively. The proposed IRT-Multiregressor All Traditional approach achieved Kappa (0.421), QWK (0.581) and Pearson correlation (0.587). These results represent improvements of 66.4\%, 8.0\%, and 1.2\% over the best ensemble models, respectively.

\begin{table*}[!htb]
\centering
\caption{Performance of the models on the Extend Essay-BR}
\label{tab:result_extend_essay}
\resizebox{\columnwidth}{!}
{%
\begin{tabular}{llccc}
\hline
\multicolumn{1}{c}{\multirow{2}{*}{\textbf{Approach}}} & \multicolumn{1}{c}{\multirow{2}{*}{\textbf{Model}}} & \multicolumn{3}{c}{\textbf{Extend Essay-BR}} \\ 
\multicolumn{1}{c}{} & \multicolumn{1}{c}{} & \textbf{Kappa} & \textbf{QWK} & \textbf{Pearson} \\ \hline
\multirow{10}{*}{Traditional} & Bayesian Ridge & 0.230 (0.018) & 0.508 (0.023) & 0.554 (0.024) \\
 & Cat Boost Regressor & 0.247 (0.016) & \textbf{0.533 (0.028)} & \textbf{0.572 (0.029)} \\
 & Decision Tree Regressor & 0.113 (0.020) & 0.352 (0.029) & 0.352 (0.029) \\
 & Extra Trees Regressor & 0.218 (0.021) & 0.492 (0.030) & 0.551 (0.029) \\
 & LGBM Regressor & 0.236 (0.018) & 0.525 (0.022) & 0.560 (0.024) \\
 & Linear Regression & 0.248 (0.015) & 0.525 (0.019) & 0.559 (0.019) \\
 & MLP Regressor & 0.173 (0.030) & 0.490 (0.036) & 0.502 (0.041) \\
 & Random Forest & 0.217 (0.019) & 0.489 (0.024) & 0.545 (0.022) \\
 & SVR & \textbf{0.250 (0.019)} & 0.528 (0.022) & 0.569 (0.023) \\
 & XGB Regressor & 0.209 (0.021) & 0.506 (0.022) & 0.527 (0.021) \\ \hline
\multirow{2}{*}{BERT} & Base & \textbf{0.343 (0.015)} & \textbf{0.641 (0.021)} & \textbf{0.672 (0.015)} \\
 & Distilbert & 0.328 (0.022) & 0.623 (0.028) & 0.641 (0.020) \\ \hline
\multirow{3}{*}{Ensemble} & Stacked Linear Regression & \textbf{0.253 (0.018)} & \textbf{0.538 (0.019)} & \textbf{0.580 (0.019)} \\
 & Stacked SVR & 0.233 (0.021) & 0.518 (0.021) & 0.554 (0.020) \\
 & Voting Regressor & 0.246 (0.012) & 0.524 (0.018) & 0.570 (0.019) \\ \hline
\multirow{3}{*}{Proposed} & All Traditional & 0.421 (0.019) & 0.581 (0.021) & 0.587 (0.020) \\
 & All BERT & \textbf{0.516† (0.013)} & \textbf{0.656† (0.014)} & \textbf{0.674† (0.014)} \\
 & All Traditional and BERT & 0.439 (0.019) & 0.597 (0.020) & 0.602 (0.020) \\ \hline
\end{tabular}
}
\end{table*}

As seen in Table \ref{tab:result_extend_essay}, BERT-based models obtained better scores than traditional models. The BERT-Base model obtained the highest scores for Kappa (0.343), QWK (0.641) and Pearson correlation (0.672). DistilBERT also outperformed traditional models but had lower results than BERT-Base. The IRT-Multiregressor All BERT approach achieved 0.516, 0.656, and 0.674, surpassing BERT-Base by 50.4\% in Kappa, 2.3\% in QWK, and 0.3\% in Pearson correlation. In another analysis of EEBR, the All Traditional and BERT approach achieved Kappa (0.439), QWK (0.597), and Pearson correlation (0.602). Compared to All Traditional, there was an increase of 4.3\%, 2.7\%, and 2.6\%, respectively. When compared to All BERT, the approach showed a reduction of 14.9\%, 9.0\%, and 10.7\%. The results of IRT-Multiregressor All BERT indicate that combining BERT models with IRT adjustment improves score evaluation.

Table \ref{tab:result_brazilian_essay} presents the performance of different approaches evaluated on the second dataset BPNE dataset. Among the traditional models, the highest values were obtained by LGBM Regressor for Kappa (0.300), SVR for QWK (0.504), and Bayesian Ridge, LGBM Regressor, and SVR for Pearson correlation (0.525). The ensemble approaches showed variations in performance, with Stacked Linear Regression achieving the highest results, reaching 0.282, 0.472, and 0.515, respectively. The IRT-Multiregressor All Traditional approach obtained 0.347, 0.450, and 0.456, respectively. Compared to the best traditional and ensemble models, this approach improved Kappa by 15.7\% and 23.0\% but had a reduction of 10.7\% and 4.7\% in QWK and 13.1\% and 11.4\% in Pearson correlation, respectively.

\begin{table*}[!htb]
\centering
\caption{Performance of the models on the Brazilian Portuguese Narrative Essays}
\label{tab:result_brazilian_essay}
\resizebox{\columnwidth}{!}
{%
\begin{tabular}{llccc}
\hline
\multicolumn{1}{c}{\multirow{2}{*}{\textbf{Approach}}} & \multicolumn{1}{c}{\multirow{2}{*}{\textbf{Model}}} & \multicolumn{3}{c}{\textbf{Brazilian Portuguese Narrative Essays}} \\
\multicolumn{1}{c}{} & \multicolumn{1}{c}{} & \textbf{Kappa} & \textbf{QWK} & \textbf{Pearson} \\ \hline
\multirow{10}{*}{Traditional} & Bayesian Ridge & 0.295 (0.085) & 0.482 (0.069) & \textbf{0.525 (0.070)} \\
 & Cat Boost Regressor & 0.253 (0.056) & 0.451 (0.053) & 0.487 (0.049) \\
 & Decision Tree Regressor & 0.189 (0.066) & 0.359 (0.070) & 0.362 (0.070) \\
 & Extra Trees Regressor & 0.226 (0.057) & 0.429 (0.047) & 0.461 (0.045) \\
 & LGBM Regressor & \textbf{0.300 (0.057)} & 0.502 (0.054) & \textbf{0.525 (0.052)} \\
 & Linear Regression & 0.213 (0.064) & 0.441 (0.046) & 0.444 (0.047) \\
 & MLP Regressor & 0.273 (0.075) & 0.478 (0.069) & 0.491 (0.070) \\
 & Random Forest & 0.284 (0.064) & 0.470 (0.050) & 0.517 (0.051) \\
 & SVR & 0.279 (0.067) & \textbf{0.504 (0.052)} & \textbf{0.525 (0.052)} \\
 & XGB Regressor & 0.269 (0.073) & 0.484 (0.046) & 0.503 (0.047) \\ \hline
\multirow{2}{*}{BERT} & Base & \textbf{0.354 (0.088)} & \textbf{0.547 (0.071)} & \textbf{0.584† (0.066)} \\
 & Distilbert & 0.300 (0.090) & 0.498 (0.072) & 0.535 (0.060) \\ \hline
\multirow{3}{*}{Ensemble} & Stacked Linear Regression & \textbf{0.282 (0.067)} & \textbf{0.472 (0.057)} & \textbf{0.515 (0.054)} \\
 & Stacked SVR & 0.135 (0.143) & 0.233 (0.238) & 0.239 (0.245) \\
 & Voting Regressor & 0.264 (0.065) & 0.459 (0.056) & 0.494 (0.052) \\ \hline
\multirow{3}{*}{Proposed} & All Traditional & 0.347 (0.057) & 0.450 (0.060) & 0.456 (0.059) \\
 & All BERT & \textbf{0.439† (0.082)} & \textbf{0.563† (0.071)} & \textbf{0.584† (0.064)} \\
 & All Traditional and BERT & 0.357 (0.065) & 0.460 (0.064) & 0.466 (0.063) \\ \hline
\end{tabular}
}
\end{table*}

As seen in Table \ref{tab:result_brazilian_essay}, BERT-based models obtained better scores than traditional models. The BERT-Base model obtained the highest scores for Kappa (0.354), QWK (0.547), and Pearson correlation (0.584). The IRT-Multiregressor All BERT approach, which combined BERT models with IRT adjustments, reached 0.439, 0.563, and 0.584, respectively. This approach surpassed BERT-Base by 24.0\%, 2.9\%, and maintained the same Pearson correlation. In a final analysis of BPNE dataset, the All Traditional and BERT approach achieved Kappa (0.357), QWK (0.460), and Pearson correlation (0.466). Compared to All Traditional, there was an increase of 2.9\%, 2.2\%, and 2.2\% in the same metrics. When compared to All BERT, the approach showed a reduction of 18.7\%, 18.3\%, and 20.2\%. These results indicate that combining multiple BERT models with IRT adjustments improved performance across all metrics.

\section{Discussions, Limitations and Directions for Future Research}\label{sec:discussion}

Cohesion is considered one of the most important aspects of well-written essays, facilitating the establishment of meaningful relationships between text components to ensure fluency and comprehensibility \citep{halliday1976cohesion,antunes2005lutar,koch2010coesao}. The demand for automated methods to assess cohesion has driven extensive research into computational approaches \citep{crossley2019tool, lima2018automatic}, yet existing models often struggle with inconsistencies between feature-based predictions and human evaluations \citep{oliveira2022estimando, oliveira2023towards}. This study advances the field by introducing an IRT-based multiregressor approach that improves cohesion score predictions while addressing these inconsistencies.

\subsection{IRT-Multiregressor Contributions}

The IRT-Multiregressor approach demonstrated superior performance in estimating cohesion scores across key evaluation metrics, including Kappa, Quadratic Weighted Kappa, and Pearson correlation, outperforming traditional and ensemble models. By integrating traditional machine learning methods with BERT-based models, the IRT framework leverages the strengths of contextual embeddings while accounting for prediction variability through its error expectation mechanism \citep{ferreira2023beta}. This combination enabled the model to capture nuanced patterns in essay cohesion that are often missed by conventional approaches.

For instance, the IRT model effectively integrates diverse feature sets, allowing for a more robust and interpretable prediction process. Unlike standard ensemble techniques, which can suffer from diminishing returns when combining overlapping features, the IRT-based integration aligns model outputs with the probability of error, producing predictions that are both precise and reliable. These technical advancements not only improve prediction accuracy but also pave the way for more explainable and user-friendly automated scoring systems.

\subsection{Broader Educational Implications}

\subsubsection{Personalised Feedback and Self-Regulated Learning (SRL)}

The ability to accurately assess cohesion opens up opportunities for personalised feedback systems. For example, by identifying specific deficiencies such as limited lexical diversity or inconsistent use of referential cohesion, educators could use the proposed model to recommend targeted exercises to students \citep{pan2025exploring}. A practical scenario might involve the model flagging a student’s weak performance in referential cohesion and suggesting activities focused on pronoun usage to enhance textual connectivity.

Additionally, integrating this approach into dashboards could provide students with actionable performance indicators (e.g., ``Improve lexical diversity by 10\%''). This aligns with SRL principles, empowering students to self-assess and address their weaknesses, while also promoting autonomy and reducing dependency on automated tools \cite{de2025development}.

\subsubsection{Curriculum Design and Data-Driven Pedagogy}

The systematic analysis enabled by the proposed model can offer critical insights into common gaps in students’ writing skills. For instance, if the model consistently identifies sequential cohesion as a challenge for high school essays, educators can adapt curricula to include activities that focus on temporal and logical connectives \citep{pan2025exploring, a2024towards}. These insights allow for data-driven pedagogical interventions that directly address students’ needs, ensuring more effective teaching strategies.

Moreover, the proposed IRT framework can support large-scale analyses to inform educational policy. For example, trends in cohesion performance across schools or regions could highlight systemic issues, enabling targeted support and resource allocation.

\subsubsection{Ethical and Explainable AI in Educational Analytics}
Transparency in automated essay scoring is essential for fostering trust and adoption among educators and students. The proposed model can leverage explainable AI (xAI) techniques such as SHAP and LIME to highlight key factors influencing cohesion scores, such as insufficient use of connectives or limited syntactic diversity. Evaluative AI approaches could further enhance this transparency by presenting multiple hypotheses for low scores, enabling a nuanced understanding of writing challenges \citep{xai2025book}. For example, the model might attribute low cohesion to a combination of limited lexical variety, inadequate use of pronouns, and weak sentence transitions.

Integrating this explainable framework into Learning Management Systems (LMS) like Moodle or Canvas offers significant potential for enhancing student learning. Real-time essay assessment could be coupled with generative AI tools such as ChatGPT to provide actionable, personalised feedback. While the IRT model identifies areas needing improvement, ChatGPT could interpret these insights in accessible ways, suggesting specific revisions like using synonyms to increase lexical diversity or providing examples of cohesive transitions. By making complex technical feedback clearer and more engaging, this integration empowers students to address their writing challenges effectively, while also equipping educators with transparent, data-driven tools to support instruction.

Such a system not only ensures responsible AI adoption but also reinforces the educator's role in guiding learning, mitigating overreliance on AI-generated outputs, and advancing the ethical use of analytics in education.

\subsection{Limitations and Directions for Future Research} \label{sec:limitation}

While the proposed approach demonstrates promising results, this study has several limitations. First, although we have explored a wide range of traditional machine learning algorithms, we did not explore different types of Large Language Models (LLMs), or methods related to parameter tuning and data balancing. This research was limited to evaluating ML models commonly used in the literature, which depend on low computational resources for a large-scale unplugged implementation. Therefore, in future work, we intend to explore other more sophisticated language models, such as open-source LLMs that are adapted to educational contexts. These methods could improve the outcome of the algorithms in the IRT approach. Second, the datasets used in this study primarily consisted of essays in Portuguese, limiting the evaluation of the model's generalizability. Future work should expand this approach to include other text types, additional essay scoring competencies, and corpora in different languages, including English.

Finally, this study did not assess the practical application of the proposed method in real-world educational environments. Developing a learning analytics tool to implement this approach and evaluating its usability, effectiveness, and satisfaction among educators and students could offer valuable insights. As future work, we recommend the development of a plugged or unplugged learning analytics platform that enables the automated assessment and feedback of essays for educators and students. This type of solution allows for large-scale implementation and focuses on personalised qualitative analysis of textual productions, enabling more effective pedagogical intervention related to writing. This implementation aims to provide more equitable and inclusive support for those in educational environments with limited resources.

\section{Conclusion} 
\label{sec:conclusion}

This study advances the field of automated essay scoring by proposing an innovative IRT-based approach to evaluate textual cohesion in Portuguese essays. By integrating traditional machine learning algorithms with contextual embeddings from BERT and leveraging IRT for score adjustment, the proposed method demonstrated superior performance across key evaluation metrics, including Kappa, QWK, and Pearson correlation. The results highlight the potential of IRT-based models to address prediction variability and better align computational outputs with human assessments.

Beyond its technical contributions, this work lays the foundation for several practical applications in educational contexts. The proposed model’s ability to provide granular insights into cohesion-related writing deficiencies can inform tailored feedback, support self-regulated learning, and guide data-driven pedagogical interventions. Furthermore, its integration into Learning Management Systems, combined with Generative AI tools, could transform essay assessment processes by making feedback more accessible and actionable for both students and educators.

Looking ahead, the study opens opportunities for further exploration, including the application of this approach to diverse text types, other languages, and real-world educational settings. By bridging the gap between advanced AI techniques and practical educational applications, this research contributes to the growing intersection of learning analytics, writing analytics, and explainable AI, paving the way for more effective and equitable teaching and learning practices.

\bibliographystyle{elsarticle-num-names} 
\bibliography{bibliography}






\end{document}